\documentclass[letterpaper, 10 pt, conference]{IEEEtran}
\IEEEoverridecommandlockouts
\usepackage[T1]{fontenc}
\usepackage{cite}

\ifCLASSINFOpdf
  \usepackage[pdftex]{graphicx}
  \graphicspath{{../resources/}}
  \DeclareGraphicsExtensions{.pdf,.jpeg,.png,.jpg}
\else
  \usepackage[dvips]{graphicx}
  \graphicspath{{../resources/}}
  \DeclareGraphicsExtensions{.eps}
\fi

\usepackage[utf8]{inputenc}
\usepackage{xcolor}
\usepackage{bbm}
\usepackage{amsmath}
\usepackage{amssymb}
\usepackage{amsfonts}
\usepackage{balance}
\usepackage{booktabs}
\usepackage{siunitx}
\usepackage{algorithmic}
\usepackage{dcolumn}
\usepackage{array}
\ifCLASSOPTIONcompsoc
  \usepackage[caption=false,font=normalsize,labelfont=sf,textfont=sf]{subfig}
\else
  \usepackage[caption=false,font=footnotesize]{subfig}
\fi

\usepackage{url}

\newcommand{\rpm}{\sbox0{$1$}\sbox2{$\scriptstyle\pm$}
  \raise\dimexpr(\ht0-\ht2)/2\relax\box2 }

\begin{document}

\title{\LARGE\bf Visual Navigation in Real-World Indoor Environments Using End-to-End Deep Reinforcement Learning}
\author{Joná\v{s}~Kulhánek$^{1,*}$, Erik Derner$^{2}$, and Robert Babu\v{s}ka$^{3}$
\thanks{$^{1}$Jon\'a\v{s} Kulh\'anek is with the Czech Institute of Informatics, Robotics, and Cybernetics, Czech Technical University in Prague, 16636 Prague, Czech Republic {\tt\footnotesize jonas.kulhanek@live.com}}%
\thanks{$^{2}$Erik Derner is with the Czech Institute of Informatics, Robotics, and Cybernetics, Czech Technical University in Prague, 16636 Prague, Czech Republic and with the Department of Control Engineering, Faculty of Electrical Engineering, Czech Technical University in Prague, 16627 Prague, Czech Republic {\tt\footnotesize erik.derner@cvut.cz}}%
\thanks{$^{3}$Robert Babu\v{s}ka is with Cognitive Robotics, Faculty of 3mE, Delft University of Technology, 2628 CD Delft, The Netherlands and with the Czech Institute of Informatics, Robotics, and Cybernetics, Czech Technical University in Prague, 16636 Prague, Czech Republic {\tt\footnotesize r.babuska@tudelft.nl}}%
\thanks{$^{*}$ Corresponding author}%
}
\maketitle

\begin{abstract}
Visual navigation is essential for many applications in robotics, from manipulation, through mobile robotics to automated driving. Deep reinforcement learning (DRL) provides an elegant map-free approach integrating image processing, localization, and planning in one module, which can be trained and therefore optimized for a given environment. However, to date, DRL-based visual navigation was validated exclusively in simulation, where the simulator provides information that is not available in the real world, e.g., the robot's position or image segmentation masks. This precludes the use of the learned policy on a real robot. 
Therefore, we propose a novel approach that enables a direct deployment of the trained policy on real robots. We have designed visual auxiliary tasks, a tailored reward scheme, and a new powerful simulator to facilitate domain randomization. The policy is fine-tuned on images collected from real-world environments.
We have evaluated the method on a mobile robot in a real office environment. Training took \textasciitilde{}30 hours on a single GPU. In 30 navigation experiments, the robot reached a 0.3-meter neighborhood of the goal in more than 86.7\,\% of cases.
This result makes the proposed method directly applicable to tasks like mobile manipulation.
%
\end{abstract}


\vspace{0.2cm}

%
\IEEEpeerreviewmaketitle
\section{Introduction}
Vision-based navigation is essential for a broad range of robotic applications, from industrial and service robotics to automated driving. The wide-spread use of this technique will be further stimulated by the availability of low-cost cameras and high-performance computing hardware. 

Conventional vision-based navigation methods usually build a map of the environment and then use planning to reach the goal. They often rely on precise, high-quality stereo cameras and additional sensors, such as laser rangefinders, and generally are computationally demanding. As an alternative, end-to-end deep-learning systems can be used that do not employ any map. They integrate image processing, localization, and planning in one module or agent, which can be trained and therefore optimized for a given environment. While the training is computationally demanding, the execution of the eventual agent's policy is computationally cheap and can be executed in real time (sampling times \textasciitilde50\,ms), even on light-weight embedded hardware, such as NVIDIA Jetson. These methods also do not require any expensive cameras.

The successes of deep reinforcement learning (DRL) on game domains \cite{mnih2013,badia2020,openai2019} inspired the use of DRL in visual navigation. 
As current DRL methods require many training samples, it is impossible to train the agent directly in a real-world environment. Instead, a simulator provides the training data \cite{wu2018,zhu2017,kulhanek2019,jaderberg2016,wu2019vn,yang2018,shah2018,ye2018,devo2020,chen2019,mirowski2018}, and domain randomization \cite{tobin2017} is used to cope with the simulator-reality gap.
However, there are several unsolved problems associated with the above simulator-based methods:
\begin{itemize}
    \item The simulator often provides the agent with features that are not available in the real world: the segmentation masks \cite{wu2018,kulhanek2019,shah2018}, distance to the goal, stopping signal \cite{wu2018,zhu2017, jaderberg2016,kulhanek2019,mirowski2018,ye2018,devo2020}, etc., either as one of the agent's inputs \cite{wu2018,shah2018} or in the form of an auxiliary task \cite{kulhanek2019}. While this improves the learning performance, it precludes a straightforward transfer from simulation-based training to real-world deployment. For auxiliary tasks using segmentation masks during training \cite{kulhanek2019}, another deep neural network could be used to annotate the inputs \cite{hu2018,tan2019}. However, this would introduce additional overhead and noise to the process and diminish the performance gain.
    \item Another major problem with the current approaches \cite{wu2018,zhu2017,jaderberg2016,kulhanek2019,mirowski2018,ye2018,devo2020} is that during the evaluation, the agent uses yet other forms of input provided by the simulator. In particular, it relies on the simulator to terminate the navigation as soon as the agent gets close to the goal. Without this signal from the simulator, the agent never stops, and after reaching its goal, it continues exploring the environment. If we provide the agent with the termination signal during training, in some cases, the agent learns an efficient random policy, ignores the navigation goal, and tries only to explore the environment efficiently.
\end{itemize}

To address these issues, we propose a novel method for DRL visual navigation in the real world, with the following contributions:
\begin{enumerate}
\item We designed a set of auxiliary tasks that do not require any input that is not readily available to the robot in the real world. Therefore, we can use the auxiliary tasks both during the pre-training on the simulator and during the fine-tuning on previously unseen images collected from the real-world environment.

\item Similarly to \cite{yang2018,wu2019vn}, we use an improved training procedure that forces the agent to automatically detect whether it reached the goal. This enables the direct use of the trained policy in the real world where the information from the simulator is not available.

\item To make our agent more robust and to improve the training performance, we have designed a fast and realistic environment simulator based on Quake III Arena \cite{connors1999} and DeepMind Lab \cite{beattie2016}. It allows for pre-training the models quickly, and thanks to the high degree of variation in the simulated environments, it helps to bridge the reality gap. Furthermore, we propose a procedure to fine-tune the pre-trained model on a dataset consisting of images collected from the real-world environment. This enables us to use a lot of synthetic data collected from the simulator and still train on data visually similar to the target real-world environment.

\item  To demonstrate the viability of our approach, we report a set of experiments in a real-world office environment, where the agent was trained to find its goal given by an image. 

\end{enumerate}

\section{Related work}


The application of DRL to the visual navigation domain has attracted a lot of attention recently \cite{wu2018,zhu2017,kulhanek2019,jaderberg2016,wu2019vn,yang2018,shah2018,ye2018,devo2020,chen2019,mirowski2018}. This is possible thanks to the deep learning successes in the computer vision \cite{tan2019,he2016} and gaming domains \cite{mnih2016,badia2020,openai2019}. However, most of the research was done in simulated environments \cite{wu2018,zhu2017,kulhanek2019,jaderberg2016,wu2019vn,yang2018,shah2018,chen2019,mirowski2018}, where one can sample an almost arbitrary number of trajectories. The early methods used a single fixed goal as the navigation target \cite{jaderberg2016}. More recent approaches are able to separate the goal from the navigation task and pass the goal as the input either in the form of an embedding \cite{wu2018,mirowski2018,yang2018} or an input image \cite{zhu2017,kulhanek2019,ye2018,devo2020}. Visual navigation was also combined with natural language in the form of instructions for the agent \cite{shah2018}.

The use of auxiliary tasks to stabilize the training was proposed in \cite{jaderberg2016} and later applied to visual navigation \cite{kulhanek2019}. In our work, we use a similar set of auxiliary tasks, but instead of using segmentation masks to build the internal representation of the observation, we use the camera images directly to be able to apply our method to the real world without labeling the real-world images. This is built on ideas from model-based DRL \cite{wahlstrom2015}.

Unfortunately, most of the current approaches \cite{wu2018,zhu2017,kulhanek2019,jaderberg2016,wu2019vn,yang2018,shah2018,ye2018,devo2020,chen2019,mirowski2018} were validated only in simulated environments, where the simulator provides the agent with signals that are normally not available in the real-world setting, such as the distance to the goal or segmentation masks. This simplifies the problem, but at the same time, precludes the use of the learnt policy on a real robot.

There are some methods \cite{mirowski2018,devo2020,ye2018} which attempt to apply end-to-end DRL to real-world domains. In \cite{mirowski2018}, the authors trained their agent to navigate in Google Street View, which is much more photorealistic than other simulators. They, however, did not evaluate their agent on a real-world robot. In \cite{ye2018}, a mobile robot was evaluated in an environment discretized into a grid with \textasciitilde{}27 states, which is an order of magnitude lower than our method. Moreover, they did not use the visual features directly but used ResNet \cite{he2016} features instead. It makes the problem easier to learn but requires the final environment to have a lot of visual features recognizable by the trained ResNet. In our case, we do not restrict ourselves to such a requirement, and our agent is much more general. Generalization across environments is discussed in \cite{devo2020}. The authors trained the agent in domain-randomized maze-like environments and experimented with a robot in a small maze. Their agent, however, does not include any stopping action, and the evaluator is responsible for terminating the navigation. Our evaluation is performed in a realistic real-world environment, and we do not require any external signal to stop the agent.

In our work, we focus on end-to-end deep reinforcement learning. We believe that it has the potential to overcome the limitations of and have superior performance to other methods, which use deep learning or DRL as a part of their pipeline, e.g., \cite{gupta2017,bhatti2016}. We also do not consider pure obstacle avoidance methods such as \cite{regier2020,xie2017,singla2019}, or methods relying on other types of input apart from the camera images, e.g., \cite{botteghi2020,sampedro2018}.

%
%
\section{Method}\label{sec:method}
We modify and extend our method \cite{kulhanek2019} to adapt it to real-world environments. We design a powerful environment simulator that uses synthetic scenes to pre-train the agent. The agent is then fine-tuned on real-world images. The implementation is publicly available on GitHub\footnote{\url{https://github.com/jkulhanek/robot-visual-navigation}}, including the source code of the simulator\footnote{\url{https://github.com/jkulhanek/dmlab-vn}}.

\subsection{Network architecture \& training}

We adopt the neural network architecture from our prior work \cite{kulhanek2019}, which uses a single deep neural network as the learning agent. It outputs a probability distribution over a discrete set of allowed actions. The input to the agent is the visual output of the camera mounted on the robot, and an image of the goal, i.e., the image taken from the agent's target pose. The previous action and reward are also used as an input to the agent. 

The network is trained using the Parallel Advantage Actor-Critic (PAAC) algorithm \cite{clemente2017} with off-policy critic updates. 
The network uses a stack of convolutional layers at the bottom, Long Short-Term Memory (LSTM) \cite{hochreiter1997} in the middle, and actor and critic heads. 
To improve the training performance, the following auxiliary tasks were used in \cite{kulhanek2019}: pixel control, reward prediction, reconstruction of the depth image, and of the observation and target image segmentation masks. Each auxiliary task has its own head, and the total loss is computed as a weighted sum of the individual losses. For further details, please refer to \cite{kulhanek2019,jaderberg2016}.



\begin{figure}[htbp]
  \centering
  \includegraphics[width=\linewidth]{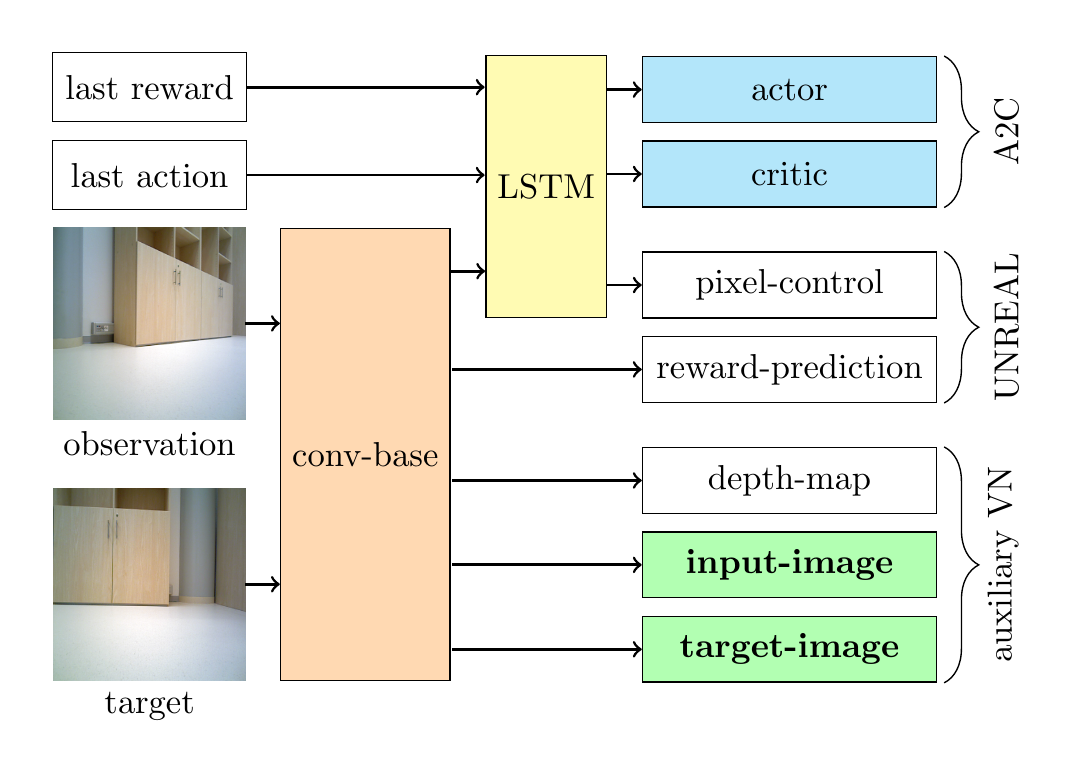}
  \caption{Neural network overview. It is similar to \cite{kulhanek2019} with the difference of the labels for VN auxiliary tasks being the raw camera images instead of the segmentation masks, which are not readily available in the real-world environment.}
  \label{fig:modules}
\end{figure}

To be able to train the agent on real-world images, we modified the set of auxiliary tasks. We no longer use the segmentation masks as they cannot be obtained from the environment without manual labeling. We have therefore replaced the two segmentation auxiliary tasks with two new auxiliary tasks: one to reconstruct the observation image and the other one to reconstruct the target image, see Fig.~\ref{fig:modules}.
This guides the convolutional base part of the network to build useful features using unsupervised learning. We hypothesized that having the auxiliary tasks share the latent-space with the actor and the critic will have a positive effect on the training performance. This hypothesis is verified empirically in Section~\ref{sec:experiments}.


Similarly to \cite{yang2018}, we have also modified the action set to include a new action called \textit{terminate}. This action stops the navigation episode and enables the trained agent to navigate in a real-world environment using only the camera images, without additional sensors to provide its actual pose in the environment. During training, when the episode terminates with the \textit{terminate} action, the agent receives either a positive or a negative reward based on its distance to the goal. The episode can also terminate (with a negative reward) after a predefined maximum number of steps.

\subsection{Environment simulator}

\begin{figure}[htbp]
    \centering
    \begin{tabular}{cc}
    \includegraphics[width=0.4\linewidth]{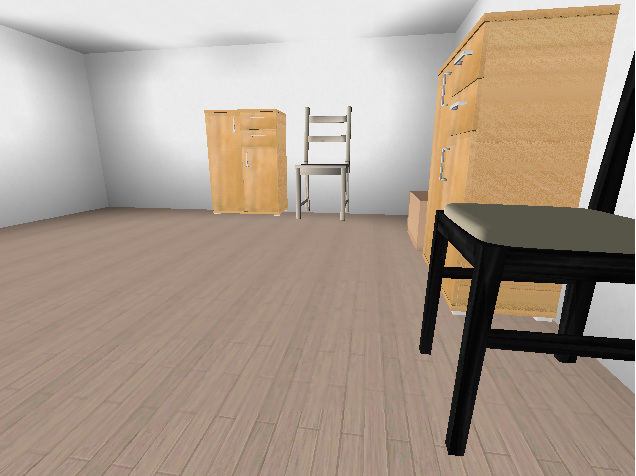} &
    \includegraphics[width=0.4\linewidth]{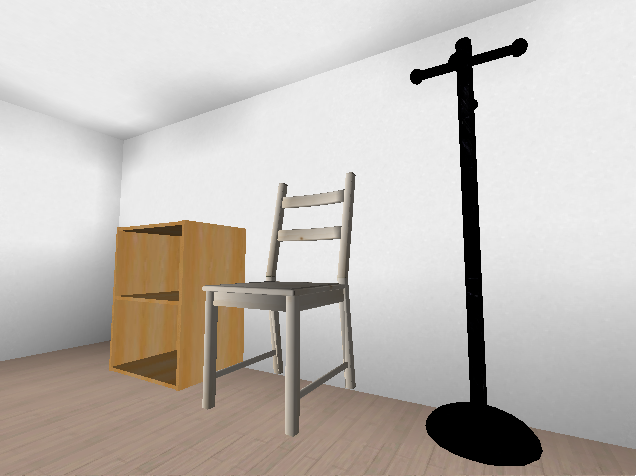} \\[0.25cm]
    \includegraphics[width=0.4\linewidth]{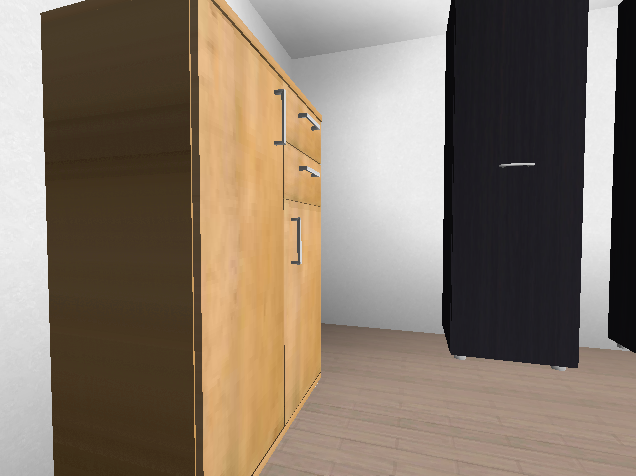} &
    \includegraphics[width=0.4\linewidth]{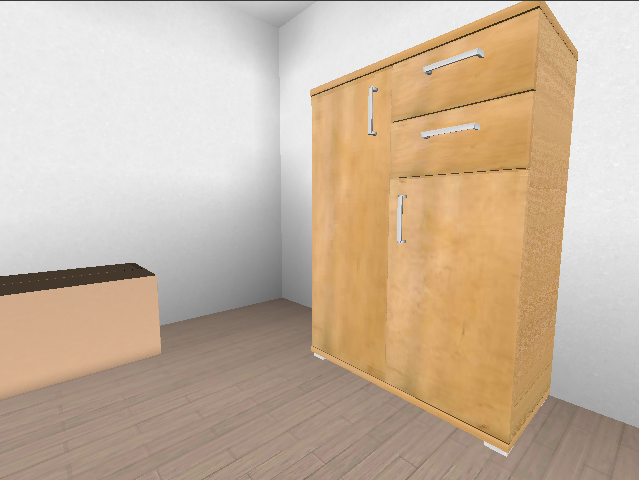}
    \end{tabular}
    \caption{Images taken from our environment simulator.}
    \label{fig:dmhouse}
\end{figure}

Since DRL requires many training samples, we first pre-train the agent in a simulated environment and then fine-tune on images collected from the real world. We have designed a novel, fast, and realistic 3D environment simulator that dynamically generates scenes of office rooms. It is based on DeepMind Lab \cite{jaderberg2016}, and it uses the Quake III Arena \cite{connors1999} rendering engine. We have extended the simulator with office-like models and textures. The room is generated by placing random objects along the walls, e.g., bookshelves, chairs, or cabinets. A target object is selected from the set of objects placed in the room, and a target image is taken from the proximity of the object. Figure~\ref{fig:dmhouse} shows four random examples of images from the environment. 
%

For each room layout, a new map is generated, which is then compiled and optimized using tools designed for Quake. To speed up the training, we keep the same room layout for 50 episodes before reshuffling the objects.

\subsection{Fine-tuning on real-world data}

Since the simulated environment does not precisely match the real-world environment, we fine-tune the agent on a set of real-world images. Throughout the rest of this paper, we restrict the motion of the robot to a rectangular grid, both for the image collection and for the final evaluation experiments.

To collect the training dataset, we placed the mobile robot equipped with an RGB-D camera in the real-world environment. We programmed the robot to automatically collect a set of images from each point of a rectangular grid, in all four cardinal directions. Odometry was used to derive and store the robot's pose (position and orientation) from which the images were taken. In this way, we obtained a dataset consisting of tuples $(x, y, \phi, i, \text{camera image}, \text{depth map})$, where $x$, $y$ and $\phi$ are the pose coordinates, and $i$ is the index of the observation image, as several images were taken from each pose.

The pre-trained agent was then fine-tuned on the collected dataset. It was initialized at a random pose, and its current policy was used to move to the subsequent pose on the grid. For each pose, a random image was sampled from the dataset, and this process repeated until the episode terminated. In this way, the agent was trained in an on-policy fashion on rollouts generated from previously collected data.

\subsection{Real-world deployment}

After the training, the agent uses its trained policy and requires only the camera observation and the target image -- no depth image or localization is necessary. To deploy our method to a new environment, we propose the following procedure. First, the agent is  pre-trained in a simulated environment using our simulator. RGB-D images are then semi-automatically collected from the real-world environment, and the agent is fine-tuned on this dataset. Finally, the agent can be placed in a real-world environment with an RGB camera as its only sensor.

%
%
\section{Experiments}\label{sec:experiments}
First, we compared several algorithms on our 3D simulated environment. Then, we fine-tuned the pre-trained model on the real-world images, and finally, we evaluated the method on the real robot.

\subsection{Implementation details}
The network architecture described in Section \ref{sec:method} was implemented as follows. The input images to the network were downsized to $84 \times 84$\,pixels. The shared convolutional part of the deep neural network had four convolutional layers. The first convolutional layer had 16 features, kernel size $8 \times 8$, and stride 4. The second convolutional layer had 32 features, kernel size $4 \times 4$, and stride 2. The third layer had 32 features with kernel size $4 \times 4$ and stride 1. The first fully-connected layer had 512 features and LSTM had also 512 output features and a single layer. The deconvolutional networks used in auxiliary tasks had two layers with kernel sizes $4 \times 4$ and strides $2$, first having 16 output features. For image auxiliary tasks, the first layer was shared. The pixel control task used a similar architecture, but it had a single fully-connected layer with $2592$ output features at the bottom, and it was dueled, as described in \cite{wang2015, jaderberg2016}. We used the discount factor $\gamma = 0.99$ for simulated experiments and $\gamma = 0.9$ for the real-world dataset. The training parameters are summarized in Table~\ref{tab:parameters}. The learning rate depends on $f$ -- the global step. Actor weight, critic weight, etc., correspond to weights of each term in the total loss \cite{kulhanek2019}.
\begin{table}[htbp]
    \caption{Method parameters.}
    \label{tab:parameters}
    \centering
    \begin{tabular}{lr}
        \toprule
        name & value \\
        \midrule
        discount factor ($\gamma$) & $0.99$ simulated / $0.9$ RW dataset \\
        maximum rollout length & 20 steps \\
        number of environment instances & 16 \\
        replay buffer size & $2\,000$ samples \\
        \midrule
        optimizer & RMSprop \\
        RMSprop alpha & $0.99$ \\
        RMSprop epsilon & $10^{-5}$ \\
        learning rate & $7 \times 10^{-4}\big(1 - \frac{f}{4 \times 10^{7}}\big)$ \\
        max. gradient norm & 0.5 \\
        \midrule
        entropy gradient weight & 0.001 \\
        actor weight & 1.0 \\
        critic weight & 0.5 \\
        off-policy critic weight & 1.0 \\
        pixel control weight & 0.05 \\
        reward prediction weight & 1.0 \\
        depth-map prediction weight & 0.1 \\
        \multicolumn{2}{l}{observation image segmentation prediction weight \hfill 0.1} \\
        target segmentation prediction weight & 0.1 \\
        pixel control discount factor & 0.9 \\
        pixel control downsize factor & 4 \\
        auxiliary VN downsize factor & 4 \\
        \bottomrule
    \end{tabular}
\end{table}

\subsection{Experiment configuration}

For both the simulated environment and the real-world dataset, we have evaluated the performance of the proposed algorithm and compared it to the performance of relevant baseline algorithms. The results were computed from $1\,000$ runs where the starting position and the goal were sampled randomly. We present the mean cumulative reward and the mean success rate -- the percentage of cases when the agent reached its goal.

\subsection{Simulated environment}
\begin{figure}[htbp]
  \centering
 \includegraphics[width=\linewidth]{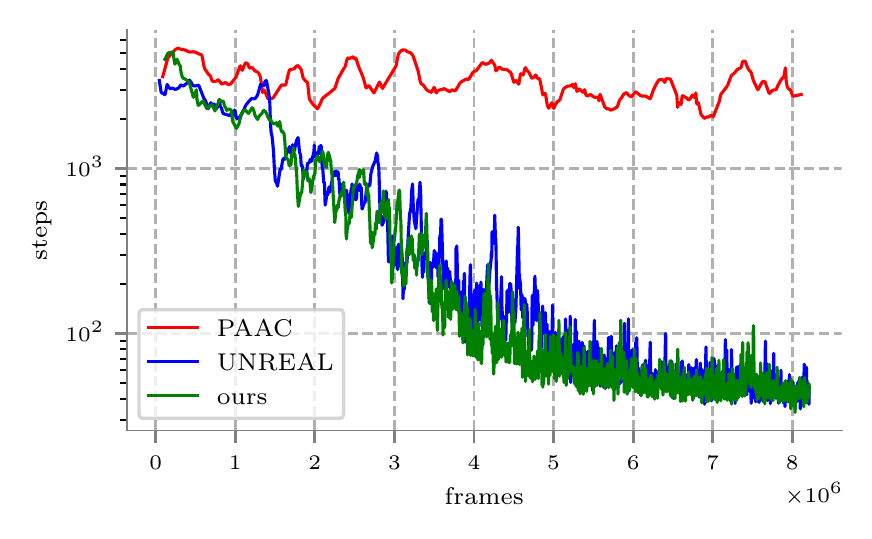}
 \caption{The plot shows the average episode length during training in the simulated environment. The PAAC, UNREAL, and our algorithm are compared. Note that we plot the average episode length, which may not correlate with the success rate -- the probability of signaling the goal correctly. In the case of UNREAL and our algorithm, however, this metric better shows the asymptotic behavior of the two algorithms since both converged to the success rate of one.}
  \label{fig:simulated-plot}
\end{figure}
\label{sec:training}

In the simulated environment, we trained the algorithm for $8 \times 10^6$ training frames. We gave the agent a reward of 1 if the agent reached the goal and stopped using the \textit{terminate} action. We gave it a reward of -0.1 if it used the \textit{terminate} action incorrectly close to (and looking at) an object of a different type than its goal, e.g., a bookcase while the goal was a chair. Otherwise, the reward was 0. In the case of the simulated environment, we did not stop the agent when it used the \textit{terminate} action incorrectly in order to improve the training performance.

The evaluation was done in $100$ randomly generated environments. A total of $1\,000$ simulations were evaluated with initial positions and goals randomly sampled. The mean success rate, the mean traveled distance, the mean number of steps taken, and its standard deviations are given in Table~\ref{tab:dmhouse_results}. The training performance can be seen in Figure~\ref{fig:simulated-plot}. The mean number of steps was averaged over the successful episodes only, i.e., those episodes which result in the agent signaling the goal correctly. The algorithm was compared to PAAC \cite{clemente2017}, which has comparable performance to A3C \cite{mnih2016}. We compared the proposed method also with the UNREAL method \cite{jaderberg2016}. Both alternative methods were adapted to incorporate the goal input by concatenating image channels of the target and the observation images and were trained on $8 \times 10^6$ training frames, same as our method.
%
\begin{table}[htbp]
    \caption{Method performance on simulated environments.}
    \label{tab:dmhouse_results}
    \centering
    \begin{tabular}{l c D{,}{\rpm}{-1} D{,}{\rpm}{-1}}
        \toprule
        algorithm & success rate & \multicolumn{1}{c}{distance traveled (m)} & \multicolumn{1}{c}{simulation steps} \\
        \midrule
        \textbf{ours} &  \textbf{1.000} & \textbf{7.691},\textbf{3.415} & \textbf{41.552},\textbf{25.717} \\
        PAAC & 0.420 & 66.913,30.329 & 398.214,276.642 \\
        UNREAL & 0.999 & 8.322,6.187 & 45.541,50.349 \\
        \bottomrule
    \end{tabular}
\end{table}

\subsection{Real-world dataset experiment}
\begin{figure}[htbp]
    \centering
    \includegraphics[width=0.85\linewidth]{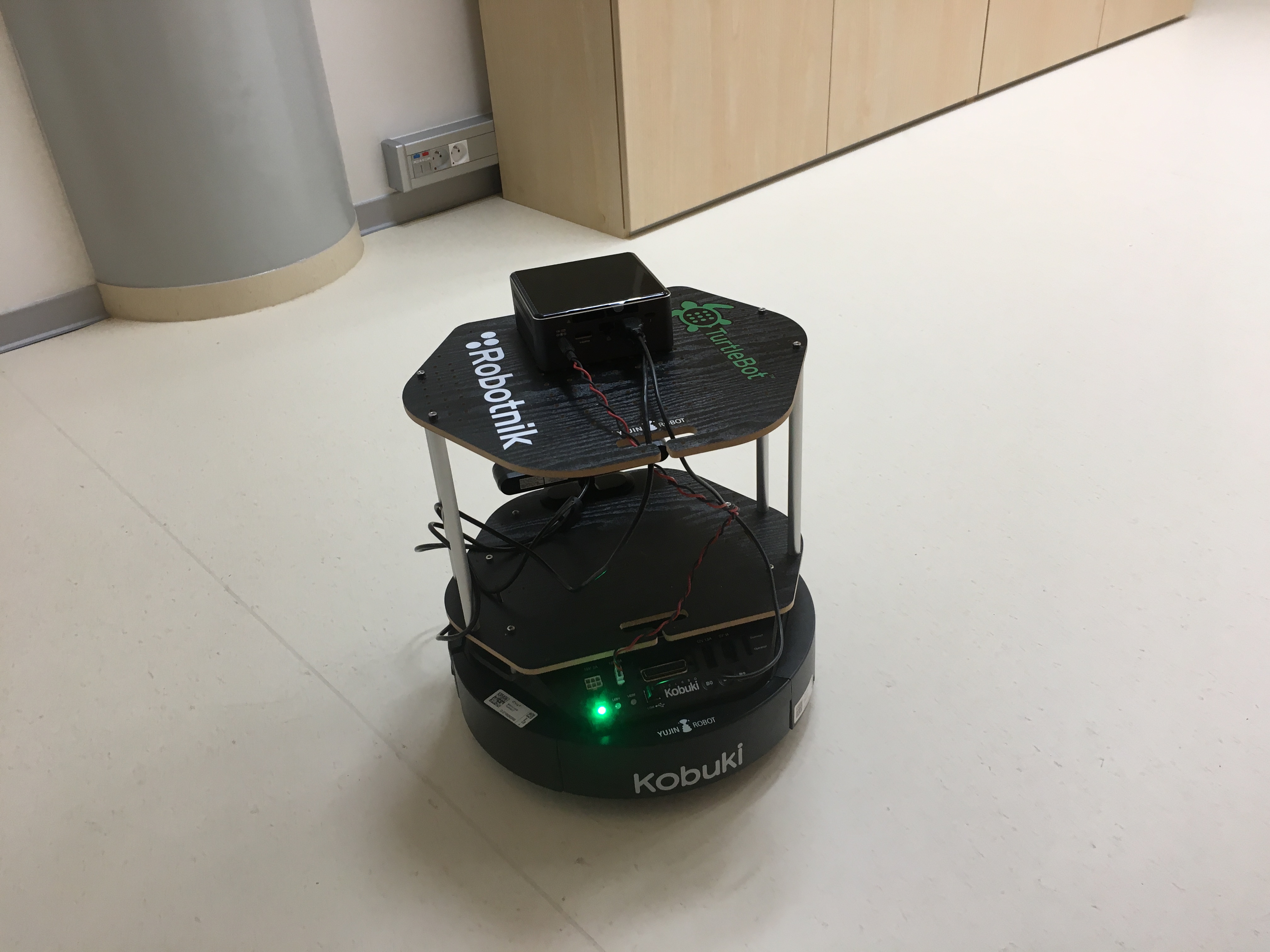}
    \caption[Turtlebot]{Mobile robot TurtleBot~2 in the experimental environment.}
    \label{fig:turtlebot}
\end{figure}
\begin{figure}[htbp]
    \centering
    \begin{tabular}{cc}
    \includegraphics[width=0.4\linewidth]{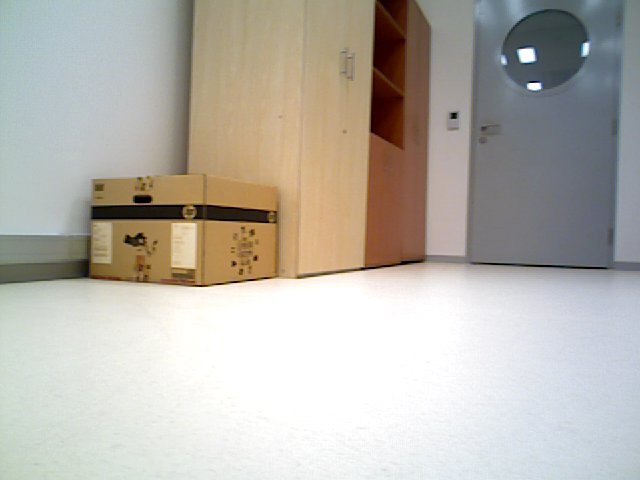} &
    \includegraphics[width=0.4\linewidth]{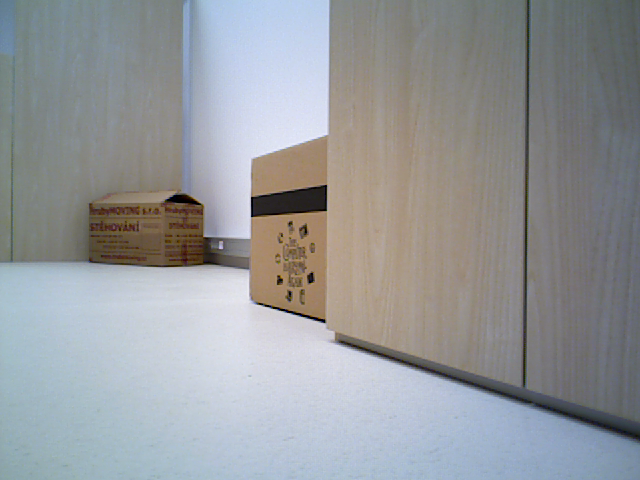} \\[0.25cm]
    \includegraphics[width=0.4\linewidth]{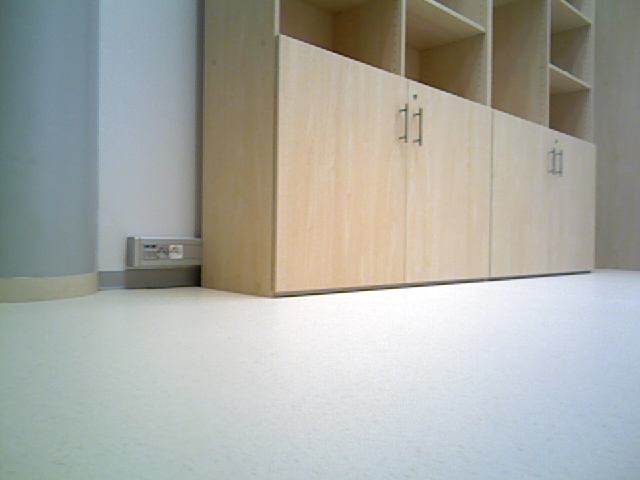} &
    \includegraphics[width=0.4\linewidth]{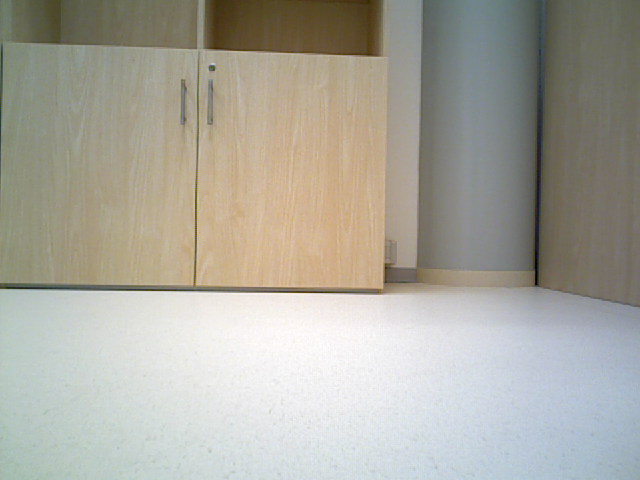}
    \end{tabular}
    \caption[Environment images]{Training images from the real-world dataset.}
    \label{fig:rw_env}
\end{figure}
\begin{table}[htbp]
  \caption{Method performance on real-world dataset.}
  \label{tab:rwsim_results}
  \centering
    \begin{tabular}{l c D{,}{\rpm}{-1} D{,}{\rpm}{-1}}
      \toprule
      algorithm & success rate & \multicolumn{1}{c}{goal distance (m)} &  \multicolumn{1}{c}{steps on grid} \\
      \midrule
      \textbf{ours} & \textbf{0}.\textbf{936} & \textbf{0.145},\textbf{0.130} & \textbf{13.489},\textbf{6.286} \\
      PAAC & 0.922 & 0.157,0.209 & 14.323,10.141 \\
      UNREAL & 0.863 & 0.174,0.173 & 14.593,9.023 \\
      \midrule
      np ours & 0.883 & 0.187,0.258 & 15.880,7.022 \\
      np PAAC & 0.860 & 0.243,0.447 & 13.699,6.065 \\
      np UNREAL & 0.832 & 0.224,0.358 & 15.676,6.578 \\
      \midrule
      random & 0.205 & 1.467,1.109 & 147.956,88.501 \\
      shortest path & -- & 0.034,0.039 & 12.595,5.743 \\
      \bottomrule
  \end{tabular}
\end{table}
\begin{figure}[htbp]
  \centering
 \includegraphics[width=\linewidth]{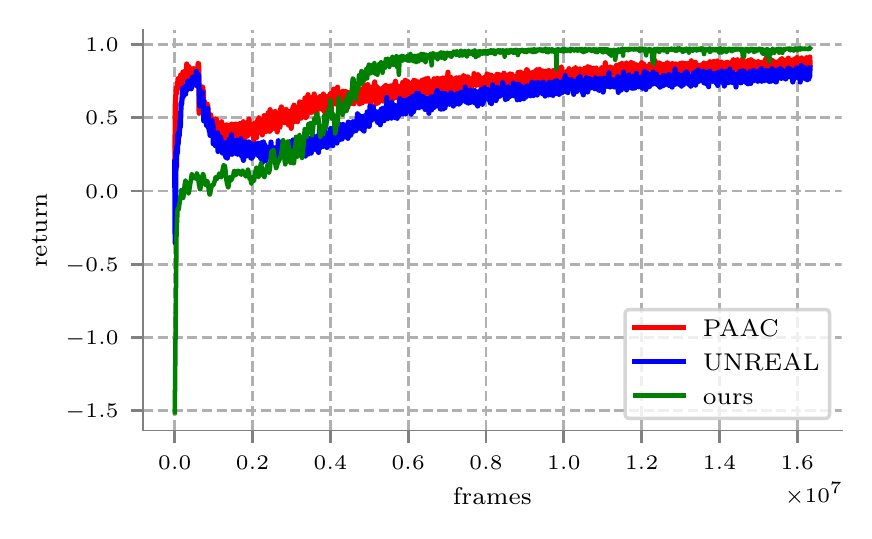}
  \caption{The plot shows the average return during training on the simulated environment.}
  \label{fig:rw-plot}
\end{figure}

In an office room, we used the TurtleBot~2 robot (Fig.~\ref{fig:turtlebot}) to collect a dataset of images taken at grid points with a $0.2\,\text{m}$ resolution. Examples of these images can be seen in Fig.~\ref{fig:rw_env}. When we collected the dataset, we estimated the robot pose through odometry. The odometry was also used for the evaluation of our trained agent in the experimental environment to assess whether the agent fulfilled its task and stopped close to the goal.

The target was sampled from a set of robot positions near a wall or near an object and facing the wall or the object. The initial position was uniformly sampled from the set of non-target states, and the initial orientation was chosen randomly. The maximal length from the initial state to the goal started at approximately three actions, and between the frames $0.5\,\times\,10^6$ and $5\,\times\,10^6$, it was increased to the maximal length, using curriculum learning \cite{kulhanek2019,mirowski2018}.

The final position of the robot was considered correct when its Euclidean distance from the target position was at most $0.3\,\text{m}$, and the difference between the robot orientation and the target orientation was at most $30^\circ$. The tolerance was set to compensate for the odometry inaccuracy. For the purpose of training, we had chosen the reward to be equal to 1 when the agent reached and stopped at its target. We penalized the agent with a reward of $-0.01$ if it tried to move in a direction that would cause a collision. Otherwise, the reward was zero.

The algorithm was trained on $3\,\times\,10^7$ frames. The training performance can be seen in Figure~\ref{fig:rw-plot}. Our method was compared to the PAAC algorithm \cite{clemente2017} and the UNREAL algorithm \cite{jaderberg2016}, modified in the same way as described in Section~\ref{sec:training}. All models were pre-trained in the simulated environment. We evaluated the algorithms on $1\,000$ simulations in total with initial positions and goals randomly sampled. The mean success rate, the mean distance from the goal (goal distance) when the goal was signaled, and the mean number of steps are given in Table~\ref{tab:rwsim_results}. We also show a comparison with the same models trained from scratch (labels beginning with \textit{np}, which stands for no pre-training) in the same table. Same as before, the mean number of steps was averaged over the successful episodes only. For comparison, we also report the shortest path and the performance of a random agent. It selects random movements, but when it reaches the target, the ground truth information is used to signal the goal.

\subsection{Real-world evaluation}
\begin{table}[htbp]
  \caption{Real-world experiment results.}
  \label{tab:rw_results}
  \centering
  \begin{tabular}{l c D{,}{\rpm}{-1} D{,}{\rpm}{-1}}
      \toprule
      evaluation & success rate & \multicolumn{1}{c}{goal distance (m)} & \multicolumn{1}{c}{steps on grid} \\
      \midrule
      mobile robot & 0.867 & 0.175,0.101 & 15.153,6.455 \\
      real-world dataset & 0.933 &  0.113,0.109 & 14.750,6.583 \\
      \bottomrule
  \end{tabular}
\end{table}
Finally, to evaluate the trained network in the real-world environment, we have randomly chosen 30 pairs of initial and target states. The trained robot was placed in an initial pose, and it was given a target image. Same as before, we show the mean number of steps, the mean distance from the goal (goal distance), and the mean success rate, where the mean number of steps was averaged over the successful episodes only. The results are summarized in Table~\ref{tab:rw_results}, where we also include the evaluation of the trained model on the real-world image dataset. The results for this dataset slightly differ from the results reported in Table~\ref{tab:rwsim_results}, as in this case, a smaller set of initial-goal state pairs was used.

\section{Discussion}

\subsection{Alternative agent objective}
In this work, we framed the navigation problem as the ability of the agent to reach the goal and stop there. To this end, we introduced the \textit{terminate} action and trained the agent to use it. Alternatively, we could stick to the approach of previous visual navigation work \cite{kulhanek2019,wu2018,wu2019vn,devo2020,jaderberg2016} in which the agent is stopped by the simulator. In the real-world environment, a localization method \cite{lowry2016,chen2017} would then have to be used to detect whether the agent reached the goal. However, this would introduce unnecessary computational overhead for the robot, and it would obscure the evaluation, as a navigation failure could be caused by either of the two systems.

\subsection{Simulated environments}
Training the agent on the simulated environments was difficult since the agent was supposed to generalize across different room layouts. In this case, the agent's goal was not to navigate to a given target by using the shortest path but to locate the target object in the environment first and then to navigate to it, possibly minimizing the total number of steps. From some initial positions, the target could not be seen directly, and the agent had to move around to see the target. Also, sometimes, there were many instances of the same object, complicating further the task.

Our proposed method achieved the best results of all compared methods, see Tab.~\ref{tab:dmhouse_results}, closely followed by the UNREAL algorithm \cite{jaderberg2016}. This clearly indicates the positive effect of using auxiliary tasks for continuous spaces. The similarity between our method and the UNREAL algorithm can be that both were close to the optimal solution and could not improve further. This is implied by the fact that both methods reached and signaled the goal successfully almost always.

\subsection{Real-world dataset}
In the case of the real-world dataset (Table~\ref{tab:rwsim_results}), the goal was to learn a robust navigation policy in a noisy environment. The noisiness came from the fact that we did not have access to the precise robot positions and orientations when the dataset was collected, but only to their estimates based on odometry, and the images were aligned to the grid points. Our method achieved the best performance. It was followed by the raw PAAC algorithm \cite{clemente2017}. The UNREAL method \cite{jaderberg2016} was the worst of these methods. We see that for all algorithms, the average number of steps to get to the target is very close to the optimal number of steps.

We did not require the agent to end up precisely in the correct position, but only in its proximity -- 30\,cm from the target position. The reward was always the same, no matter how far the agent was from the target. The noisiness in the position labels in the dataset might cause the agent to stop closer to the goal than it was necessary to meet a safety margin at the cost of more steps.

A substantial performance improvement can be observed when using pre-training on the simulated environment for all three methods. Although the simulated environment used continuous space, it looked visually different, and the control was different, it was still remarkably beneficial for the overall performance. This can be seen in Figure~\ref{fig:simulated-plot}, where all methods quickly reached a relatively high score. The decrease in performance of the UNREAL method might be caused by the ineffectiveness of its auxiliary tasks. As the environment was not continuous, but the rotation actions turned the robot by 90\,$^\circ$, the observations at these rotations were non-overlapping. Therefore, the task of predicting the effect of each action became much more difficult, and instead of helping the algorithm converge faster, it was adding some noise to the gradient. 

In our method, the auxiliary tasks might help the model learn a compact representation of each state quicker and make the algorithm converge faster. We believe the proposed visual navigation auxiliary tasks have a similar effect as using autoencoders in model-based DRL \cite{wahlstrom2015}. By learning the sufficient features to reconstruct the observation, and by using these features as the input for the policy, the policy could use a higher abstraction of the input while it learns an easier task than model-based methods and thus converges faster.

\subsection{Real-world experiment}

The results of the real-world experiments indicate that our algorithm is able to navigate in the real environment well. The discrepancy in the performance between the simulation and the real-world environment can be caused by a generalization error in transferring the learned policy as well as by the errors in odometry measurements, which were used for estimating the robot position for the experiment evaluation.

\section{Conclusion \& future work}
In this paper, we proposed a deep reinforcement learning method for visual navigation in real-world settings. 
Our method is based on the Parallel Advantage Actor-Critic algorithm boosted with auxiliary tasks and curriculum learning. It was pre-trained in a simulator and fine-tuned on a dataset of images from the real-world environment.

We reported a set of experiments with a TurtleBot in an office, where the agent was trained to find a goal given by an image. In simulated scenarios, our agent outperformed strong baseline methods. We showed the importance of using auxiliary tasks and pre-training. The trained agent can easily be transferred to the real world and achieves an impressive performance. In 86.7 \% of cases, the trained agent was able to successfully navigate to its target and stop in the 0.3-meter neighborhood of the target position facing the same way as the target image with a heading angle difference of at most 30 degrees. The average distance to the goal in the case of successful navigation was 0.175 meters.

The results show that DRL presents a promising alternative to conventional visual navigation methods. Auxiliary tasks provide a powerful way to combine a large number of simulated samples with a comparatively small number of real-world ones. We believe that our method brings us closer to real-world applications of end-to-end visual navigation, so avoiding the need for expensive sensors.

In our future work, we will evaluate this approach in larger environments with continuous motion of the robot, rather than on the grid. The experiments will be performed using a motion capture system for accurate robot localization. To extend the memory of the actor, one can pursue the idea of implicit external memory in deep reinforcement learning \cite{graves2016} and transformers \cite{parisotto2019}. By using better domain randomization, a general model can be trained that will not need the robot pose data accompanying the images in the fine-tuning phase. We also plan to work with continuous action space and model the robot dynamics.

\section*{Acknowledgment}
This work was supported by the European Regional Development Fund under the project Robotics for Industry 4.0 (reg.~no. CZ.02.1.01/0.0/0.0/15\_003/0000470). Robert Babu\v{s}ka was supported by the European Union’s H2020 project Open Deep Learning Toolkit for Robotics (OpenDR) under grant agreement No 871449.

\ifCLASSOPTIONcaptionsoff
  \newpage
\fi


\bibliographystyle{IEEEtran}
\balance
\bibliography{IEEEabrv,bibliography}

%








\end{document}